\documentclass[letterpaper, 10 pt, conference]{ieeeconf}  % Comment this line out if you need a4paper

\IEEEoverridecommandlockouts                              % This command is only needed if 
\usepackage{cite}

\usepackage{amsmath,amssymb,amsfonts}
\usepackage{algorithmic}
\usepackage{graphicx}
\usepackage{textcomp}
\usepackage{xcolor}
\usepackage{hyperref}
\usepackage{cleveref}
\usepackage{siunitx}
\usepackage{float}
\usepackage{balance}

\begin{document}

\def\robotname{Zippy}

% other title options on slack
\title{\LARGE \bf
Zippy: The smallest power-autonomous bipedal robot \thanks{This work was supported in part by the National Science Foundation under Grant CMMI-2408884.}
}

% \title{Zippy: The smallest power-autonomous 
% bipedal robot
% \thanks{This work was supported in part by the National Science Foundation under Grant CMMI-2408884.}
% }

% \author{Author1, Author2, and Author3}   
\author{
    {Steven Man\textsuperscript{*1}, Soma Narita\textsuperscript{*2}, Josef Macera\textsuperscript{*2}, Naomi Oke\textsuperscript{3}, Aaron M. Johnson\textsuperscript{3}, and Sarah Bergbreiter\textsuperscript{3}}%    
    % Thanks command for affiliation at bottom left
    \thanks{\textsuperscript{*}Equal contribution}%
    \thanks{\textsuperscript{1}Robotics Institute, \textsuperscript{2}College of Engineering, \textsuperscript{3}Department of Mechanical Engineering, Carnegie Mellon University, Pittsburgh, Pennsylvania, 15213, USA, \texttt{sman2@andrew.cmu.edu}}%
}

\maketitle

\begin{abstract}
Miniaturizing legged robot platforms is challenging due to hardware limitations that constrain the number, power density, and precision of actuators at that size. By leveraging design principles of quasi-passive walking robots at any scale, stable locomotion and steering can be achieved with simple mechanisms and open-loop control. Here, we present the design and control of ``\robotname{}", the smallest self-contained bipedal walking robot at only \SI{3.6}{\centi\meter} tall. \robotname{} has rounded feet, a single motor without feedback control, and is capable of turning, skipping, and ascending steps. At its fastest pace, the robot achieves a forward walking speed of \SI{25}{\centi\meter/\second}, which is \num{10} leg lengths per second, the fastest biped robot of any size by that metric. This work explores the design and performance of the robot and compares it to similar dynamic walking robots at larger scales. 
\end{abstract}

% \begin{IEEEkeywords}
% Passive Walking, Underactuated Robots, Mechanism Design of Mobile Robots, Humanoid and Bipedal Locomotion, 
% \end{IEEEkeywords}

\section{Introduction}
% Some motivation
Small, centimeter-scale robots have the potential to excel at traversing tight spaces found in industrial facilities, natural cavities, and disaster debris, allowing for inspection and exploration tasks typically inaccessible to robots with larger footprints. These robots often use cheaper materials in smaller quantities, making them cost-effective for deploying numerous agents across a wide range of spatially constrained environments.  

% Locomotion is tough in small robots -- make it easier by adding more legs
Designing small, autonomous robots can be difficult, in part due to the challenge of locomoting over non-flat terrain effectively and efficiently. Many small robots take advantage of a multi-legged morphology which offers robust stability and superb mobility\cite{birkmeyer2009dash,haldane2015running,demario2018development,goldberg2018power,jayaram2020scaling}. These small robots are often inspired by animals and insects that can crawl quickly over diverse surfaces and uneven terrain using four or more limbs. A significant advantage of multi-legged robots is their passive stability
%The passive stability of a robot requires that its center of mass projects within the support polygon made by its supporting limbs 
\cite{gupta2017brief}, because small robots typically do not have the sensing, actuation, and control resources to provide active dynamic stability during locomotion. 
With four or more legs, such robots can easily remain stable by both having redundant leg-ground contact and a larger support polygon. However, the increased mobility and stability of multi-legged designs sometimes come at the cost of increased complexity and footprint due to mechanisms required for integration.

\begin{figure}[t]
\vspace{.5em}
\centerline{\includegraphics[width=\columnwidth]{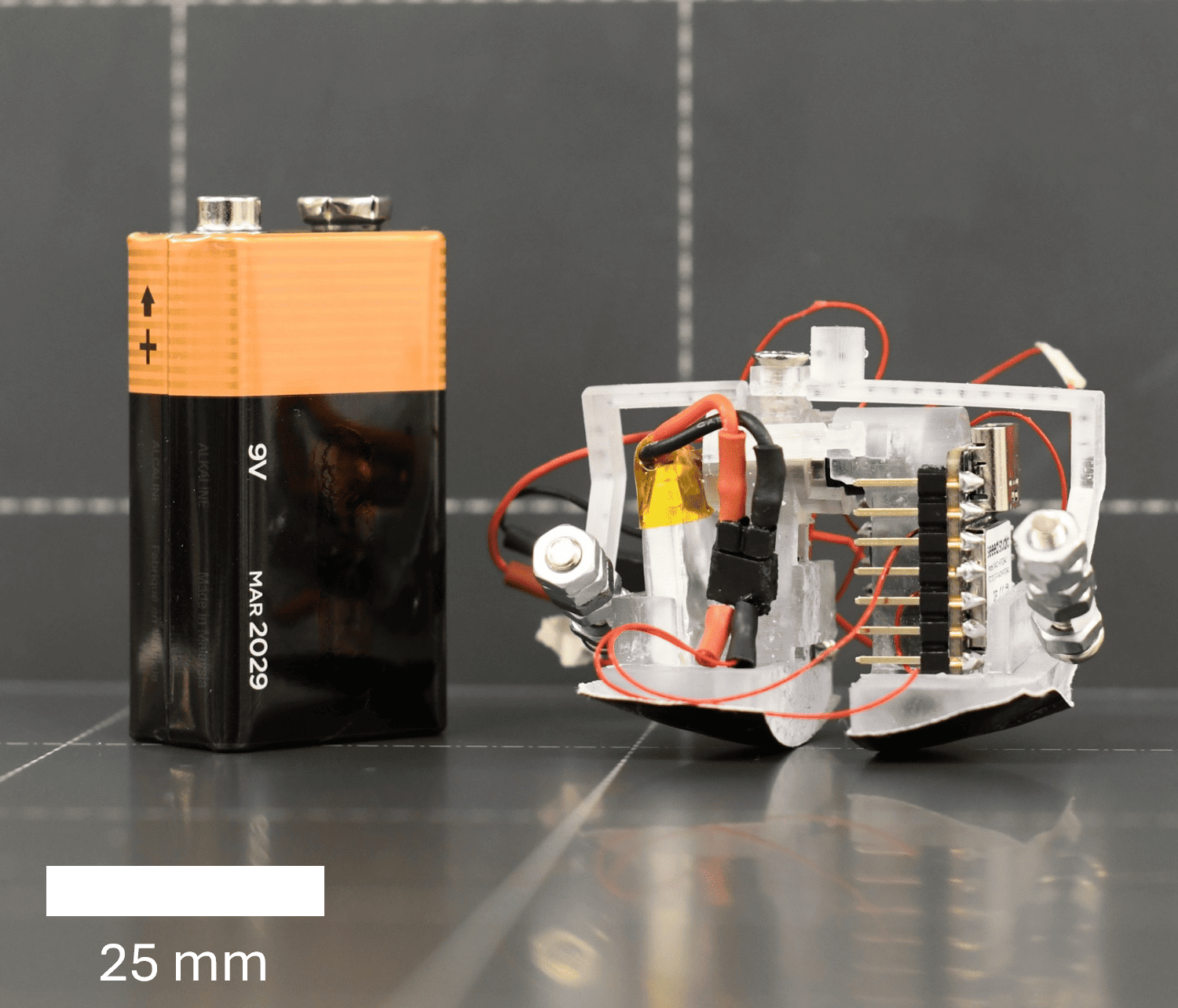}}
\caption{The smallest self-contained biped robot (right) next to a standard 9V battery (left). The robot has one actuator at the hip and rounded feet that provide stable passive dynamics.}
\label{fig:money_shot}
\end{figure}

% I can make a 2-legged robot using passive dynamics
Reducing the number of legs to two is more common at larger sizes in humanoid robot designs \cite{robots2018,gong2019feedback}. However, many of these designs require large numbers of actuators (\num{10} or more), high-precision sensors (encoders, force/torque sensors, etc), and complicated mechanisms.
A separate class of bipedal robots, passive dynamic walkers (PDWs), use the passive dynamics of the robot to achieve stability and efficiency while walking on two legs \cite{mcgeer1990passive, collins2001three}. Using only their natural dynamics, PDWs can achieve a periodic walking cycle without any actuation even in the presence of minor perturbations \cite{garcia1998simplest}.
This research has inspired powered, autonomous walking robots with noteworthy efficiency or simplicity \cite{tedrake2004actuating,collins2005bipedal, wisse2005keep, bhounsule2014low}. 

To generate the relevant passive dynamics, these robots often include key design features such as curved feet and careful placement of the hip joint relative to the center of mass \cite{collins2005efficient, kyle2023simplest, islam2022scalable, tedrake2004actuating}. These design features are the realizations of models like the rimless wheel \cite{mcgeer1989wobbling, mcgeer1990passive}, compass gait \cite{garcia1998simplest, alexander1995simple, goswami1997limit}, and other models with curved feet \cite{kuo1999stabilization, kuo2007six}.
Validating these PDW models across scales with physical hardware can further our understanding of how dynamic walking may be applied to systems at various sizes to take advantage of the simplicity and efficiency offered by passive dynamic bipedal robots.

To this end, we present \robotname{} (\Cref{fig:money_shot}), a \SI{3.6}{\centi\meter}-tall quasi-passive bipedal walking robot using a single motor that can: passively stand without any actuation, self-start from a standstill, stably locomote at speeds as fast as \num{10} leg lengths per second (the highest of any bipedal robot), achieve a flight phase during locomotion, turn left or right at a controlled rate, traverse steps and non-smooth terrain, and is self-contained with an on-board power and control system. 

\robotname{} is similar in design to Mugatu \cite{kyle2023simplest}, with rounded feet and a single actuator at the hip, but \num{1}/\SI{6}{th} the size and with some notable design changes discussed herein.  
The challenge of dynamic stability is amplified at small scales, where disturbances have larger effects. 
By overcoming this challenge, \robotname{} represents the only biped among self-contained small-scale robots (\Cref{tab:diff_small_robots}), as well as the smallest design of any dynamic bipedal robot (\Cref{tab:diff_walkers}).
It is also the fastest autonomous biped relative to its leg length (or hip height for robots with non-straight legs), more than doubling the \num{4.1} leg-lengths per second record held by Cassie \cite{crowley2023optimizing}.

\setlength{\tabcolsep}{0.31em}
\begin{table}[!tb]
\vspace{1.5em}
\caption{Comparison of small-scale power-autonomous legged robots}
\centering
\begin{tabular}{cccccc}
\multicolumn{1}{c|}{Robot} & \textbf{\robotname{}} & HAMR-F & DASH & VelociRoACH & MM3P \\ \hline
\multicolumn{1}{c|}{Citation} & \textbf{This Work} & \cite{goldberg2018power} & \cite{birkmeyer2009dash} & \cite{haldane2015running} & \cite{demario2018development}\\ 
\multicolumn{1}{c|}{Mass (\SI{}{\gram})} & \num{25.0} & \num{2.80} & \num{16.2} & \num{54.6} & \num{14.5} \\ 
\multicolumn{1}{c|}{Actuators} & \boldmath{$1$} & \num{8} & \boldmath{$1$} & \boldmath{$1$} & \boldmath{$1$} \\ 
\multicolumn{1}{c|}{Legs} & \boldmath{$2$} & \num{4} & \num{6} & \num{6} & \num{4} \\ 
\multicolumn{1}{c|}{Length (\SI{}{\milli\meter})} & \num{50.5}* & \num{45.0} & \num{100} & \num{104} & \num{47.0} \\ 
\multicolumn{1}{c|}{Speed (\SI{}{\milli\meter/\second})} & \num{250} & \num{172} & \num{1500} & \boldmath{$4900$} & \num{57.0} \\ 
\multicolumn{1}{c|}{Speed (\SI{}{BL/\second})} & \num{5.0} & \num{3.8} & \num{15} & \boldmath{$47$} & \num{1.2} \\ 
\multicolumn{1}{c|}{Minimum CoT} & \num{11.2} & \num{83.9} & \num{14.7} & \boldmath{$3.30$} & - \\ 
\multicolumn{1}{c|}{Runtime (min)} & \boldmath{$54$} & \num{4.5} & \num{40} & - & - \\ \hline

\multicolumn{5}{l}{*The robot's width is its longest dimension} \\ 
\end{tabular}
\label{tab:diff_small_robots}
\end{table}

\begin{table}[!tb]
\caption{Comparison of dynamic bipedal robots}
\centering
\begin{tabular}{cccccc}
\multicolumn{1}{c|}{Robot} & \textbf{\robotname{}} & Cassie & Mugatu & Ranger & Collins \\ \hline
\multicolumn{1}{c|}{Citation} & \textbf{This Work} & \cite{robots2018,gong2019feedback} & \cite{kyle2023simplest} & \cite{bhounsule2014low} & \cite{collins2005bipedal} \\ 
\multicolumn{1}{c|}{Mass (\SI{}{\kilo\gram})} & \boldmath{$0.025$} & \num{31} & \num{0.81} & \num{9.9} & 13 \\ 
\multicolumn{1}{c|}{Actuators} & \boldmath{$1$} & \num{10} & \boldmath{$1$} & \num{3} & 2 \\ 
\multicolumn{1}{c|}{Leg Length (\SI{}{\meter})} & \boldmath{$0.025$} & \num{1.0}* & \num{0.15} & \num{1.0} & 0.81 \\ 
\multicolumn{1}{c|}{Speed (\SI{}{\meter/\second})} & \num{0.25} & \boldmath{$4.1$} & \num{0.16} & \num{0.59} & 0.44 \\ 
\multicolumn{1}{c|}{Speed (\SI{}{LL/\second})} & \boldmath{$10$} & \num{4.1}* & \num{1.05} & \num{0.59} & 0.54 \\ 
\multicolumn{1}{c|}{Minimum CoT} & \num{11.2} & \num{0.71}* & \num{5.3} & \boldmath{$0.19$} & 0.20 \\ \hline
\multicolumn{6}{l}{*Estimated value from references} \\ 
\end{tabular}
\label{tab:diff_walkers}
\end{table}

In this work, we study the effect of design changes that enable both a high relative speed and stable locomotion over non-smooth terrain, both of which are typically challenging for passive dynamic walkers and small robots alike. We present a ``skipping gait" with a brief aerial phase that increases swing leg clearance, enabling us to re-examine how pre-existing design rules \cite{kyle2023simplest} for larger quasi-passive walkers may be modified to enhance speed and stability. We also evaluate the robot's ability to turn, performance on uneven terrain, and energy efficiency to examine its viability as a small-scale locomotion platform capable of performing meaningful tasks like inspection or search and rescue.

\section{Methods}
\subsection{Robot hardware design}
This robot builds on the design of the simplest walking robot, Mugatu, a bipedal walking robot with only a single actuator \cite{kyle2023simplest}. Mugatu established five design rules that informed the initial design of our walker.
\begin{enumerate}
    \item The center of gravity must be below the center of curvature of the foot radius. 
    \item The vertical location of the hip must be above the center of curvature of the foot radius.
    \item In the sagittal plane, the location of the hip axis must be behind the center of gravity.
    \item There must be a lateral gap between the two feet, with the curvature tangent at the floor (as first found in \cite{islam2022scalable}).
    \item The applied torque and inertia must be sufficient to overcome friction when starting.
\end{enumerate}

\begin{figure}[tbp]
\vspace{.5em}
\centerline{\includegraphics[width=0.9\columnwidth]{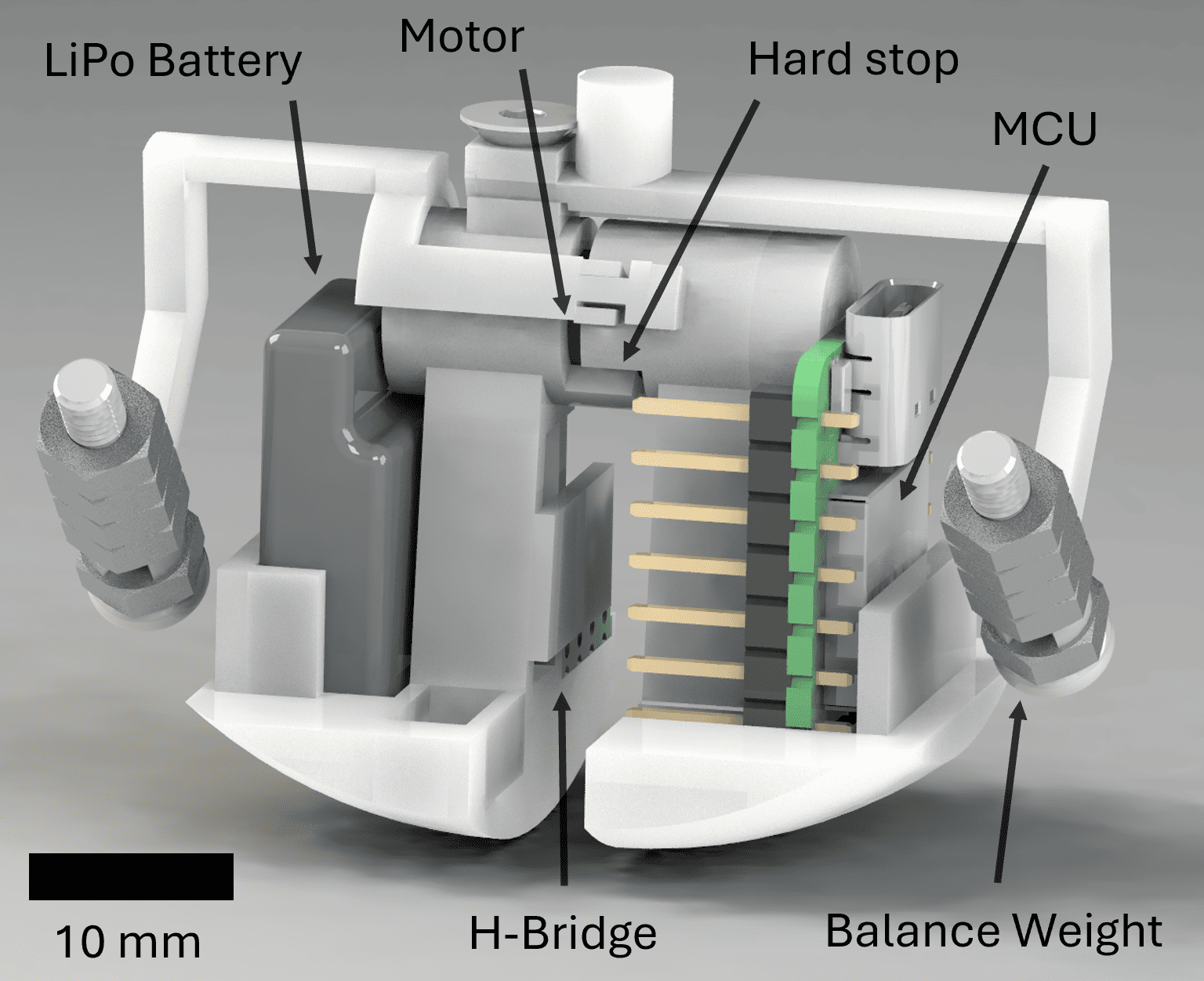}}
% \vspace{-.5em}
\caption{An annotated rendering of the \robotname{} robot design.}
\label{fig:callout}
\end{figure}

%These design rules informed the initial design of our walker.
\robotname{} consists of two rigid bodies, each being a leg connected to a curved foot and an arm (\Cref{fig:callout}). These bodies are 3D printed (Pico 2 HD, Asiga) and joined by a single open-loop DC motor at the hip (2357, Pololu). The robot uses a microcontroller board (XIAO nRF52840, Seeed Studio) to drive the motor with an H-bridge (DriveCell, Microbots) powered by a LiPo battery (\SI{3.7}{\volt}, \SI{100}{\milli\ampere\hour}). Modular fasteners act as counterbalance weights on the arms to counter yawing moments and improve the ability to lift the advancing leg, as in\cite{collins2005bipedal, kyle2023simplest}. We attached duct tape (Bomei Pack) to the bottom of the feet for an improved coefficient of friction. Modifying this contact interface helps to avoid either stick-slip, caused by insufficient friction, or stepping in place caused by excess energy loss due to high friction.

\begin{figure}[tbp]
\vspace{.5em}
\centerline{\includegraphics[width=\columnwidth]{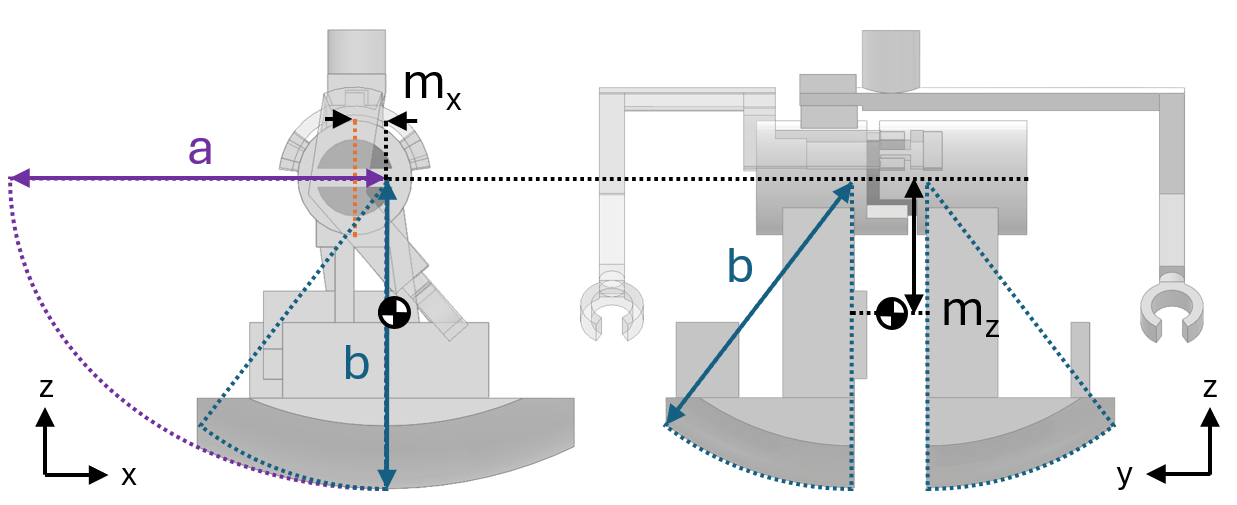}}
% \vspace{-1em}
\caption{Robot frontal and sagittal plane geometry.}
\label{fig:dims}
\end{figure}

Curved feet are the primary design feature that enables passive stability during walking and are commonly found on PDWs to aid in lateral stability \cite{kuo1999stabilization}. However, the curvature of these feet in the frontal and sagittal planes greatly affects gait generation. 
We assessed two foot shapes to ensure that the robot can stably walk agnostic of initial conditions. We first isometrically scaled the spherical feet of Mugatu \cite{kyle2023simplest} (radius $b$ in \Cref{fig:dims}) to test the resulting walking speed and stability. This configuration is referred to as ``Scaled Mugatu.'' After observing instability caused by excessive rolling and pitching during preliminary walking tests, we modified the feet to have an ellipsoidal profile ($a$ and $b$ in \Cref{fig:dims}), leading to a lower foot curvature in the sagittal plane than in the frontal plane. In this new design, we also raised the center of the ellipsoid to hip height, reducing the minimum foot curvature. The two designs were evaluated by assessing the roll and pitch amplitudes of the robot body during locomotion. The dimensions for both Scaled Mugatu and \robotname{} are provided in \Cref{tab:param_table}.

\begin{table}[t]
\centering
\caption{Robot physical parameters}
\begin{tabular}{c|cccc}
Parameter & Symbol & \robotname{} & Scaled Mugatu & Units \\ \hline
Mass & $m$ & \num{25.0} & \num{24.7} & \SI{}{\gram} \\ 
Total height & $h$ & \num{36.4} & \num{36.4} & \SI{}{\milli\meter} \\ 
Total width & $w$ & \num{50.5} & \num{50.5} & \SI{}{\milli\meter} \\ 
Foot ellipsoid major axis & $a$ & \num{30.0} & \num{19.1} & \SI{}{\milli\meter} \\ 
Foot ellipsoid minor axis& $b$ & 24.7 & 19.1 & \SI{}{\milli\meter} \\ 
CG Z offset & $z_{cg}$ & \num{10.3}* & \num{12.0}* & \SI{}{\milli\meter} \\ 
CG X offset & $x_{cg}$ & \num{3.0}* & \num{2.9}* & \SI{}{\milli\meter} \\ \hline
\multicolumn{4}{l}{*Estimated from CAD} \\
\end{tabular}
\label{tab:param_table}
\end{table}

\subsection{Robot control}
Most existing humanoid and bipedal walking robots require precise motor control, often relying on closed-loop feedback for trajectory tracking \cite{taix2013generating, sreenath2011compliant, ficht2021bipedal}.
When scaling down robot actuators, precise motor control becomes increasingly difficult due to both packaging the required sensing components onto a small platform and having to respond to the typically higher frequency disturbances at small scales.
In other passive dynamic walking robots, actuation signals are provided as motor position set points \cite{kyle2023simplest, islam2022scalable}, which rely on servo PD control to track the joint trajectories. At the scale of our walker, closed-loop servo control becomes more difficult due to the challenge in designing a small motor-encoder assembly. 

Instead, we rely on open-loop control of the robot's gait along with a physical hard stop (\Cref{fig:callout}) to constrain the range of motion. A square wave voltage is applied to the hip motor, parameterized by the amplitude, frequency, and DC offset:
\begin{equation}
V_{motor}(t) = \begin{cases} 
      A+V_0 & \text{mod}(t,\frac{1}{f}) < \frac{1}{2f} \\
      -A+V_0 & \text{mod}(t,\frac{1}{f}) \geq \frac{1}{2f} 
   \end{cases}
\label{eqn:square_wave}
\end{equation}
\noindent where $V_{motor}(t)$ is the motor voltage, $A$ is the amplitude of the square wave, $V_0$ is the DC offset of the waveform, and $f$ is the frequency. The H-bridge approximates $V_{motor}$ using pulse width modulation (PWM), and a negative $V_{motor}$ reverses the motor direction. Forward walking speed is controlled by both the amplitude and frequency, and the DC offset controls the turning radius. We transmit the desired waveform parameters and start the robot remotely via Bluetooth.

The hard stop guarantees that the hip joint stays within the desired range of motion even if the control signal would normally drive the motor past the hard stop. 
The periodic nature of our actuation scheme requires the motor to reverse its direction twice in a single period. Without the hard stop, the motor would first have to slow down to zero velocity before spinning in the opposite direction in the presence of a reversed voltage signal. The hard stop rapidly decelerates the swing leg, allowing the motor to begin its reverse motion around the same phase in sequential walking periods. %leading to more consistent steps.

We note that the primary differences between \robotname{} and Mugatu are their scale, foot shape, center of curvature, center of mass, and the inclusion of a hard stop with open-loop motor control instead of motor position control with a servo. Other parameters such as foot gap and leg length were scaled, but otherwise left unchanged.  

\subsection{Experimental setup}
\paragraph{Effect of modifying existing design rules}
To compare the difference between Scaled Mugatu and \robotname{}, we kept both robots as similar as possible except for foot curvature (\Cref{tab:param_table}). We set up cameras to record the sagittal and frontal views of locomotion at \SI{240}{fps} to analyze their respective roll and pitch oscillation amplitudes. This is the only case where Scaled Mugatu was tested. We actuated both robots with a waveform of $A = $~\SI{3.2}{\volt} and $f = $~\SI{3.7}{\hertz}.

\paragraph{Locomotion performance}
We designed three experiments to assess locomotion: forward walking speed, turning radius, and traversability of steps and rough terrain. A fixed camera (iPhone \SI{14}{} Pro, \SI{1080}{p}, \SI{30}{fps}) mounted above and leveled to the walking surface was used to record the walking speed and turning radius experiments. The camera placement enabled recording of a \SI{1}{\meter} long path.  For each trial, we ensured that the battery was charged and we start the video recording with the robot at rest next to a tape measure before remotely starting the robot. We then extract the robot's trajectory using an object-recognition and tracking software library (YOLOv8, Ultralytics). From the trajectory data, we determine the start and end points of a straight walking path, and take the average velocity of the robot across that path.

The forward walking speed experiments assessed the effect of the control parameters on the forward walking speed. We tested combinations of four voltage amplitudes ($A = $~\SI{1.75}{\volt}, \SI{2.2}{\volt}, \SI{2.75}{\volt}, \SI{3.2}{\volt}) and six stride frequencies ($f = $~\SI{2.5}{\hertz}, \SI{3.1}{\hertz}, \SI{3.7}{\hertz}, \SI{6}{\hertz}, \SI{8.3}{\hertz}, \SI{12.5}{\hertz}). % as shown in \Cref{fig:sweep}. 
The upper and lower frequency bounds were determined by stability while \SI{3.2}{\volt} was the maximum voltage that the H-bridge could supply. Outside the selected parameters, the robot exhibits chaotic walking. For each set of waveform parameters, we collected three trials of walking on a flat phenolic resin bench top and recorded their average speed.

For the robot steering experiments, we used the same camera setup to record robot walking trajectories across seven different motor voltage additional offset values (\SI{0}{\volt}, $\pm$ \SI{0.2}{\volt}, $\pm$ \SI{0.4}{\volt}, $\pm$ \SI{0.6}{\volt}) on the same surface. These experiments were performed with a waveform of $A = $~\SI{2.6}{\volt}, $f = $~\SI{3.7}{\hertz}, and $V_0 = $~\SI{0.1}{\volt} (to ensure straight nominal walking). Each offset value tested was added to this nominal offset.
Positive offset values resulted in right turns while negative offset values resulted in left turns.
We extracted the robot trajectory using the same method mentioned above and isolated the time ranges where the robot path traces an arc from \SI{0}{\degree} to \SI{90}{\degree}. We then applied a least-squares circular fit \cite{circfit} to the turning trajectory data to determine the radius of curvature as a function of the waveform offset. 

For the steps and rough terrain traversability experiments, the robot was tested using a motor voltage waveform with $f = $~\SI{3.7}{\hertz} and $A = $~\SI{3.2}{\volt} on 3D-printed steps of varying heights ($h = $~\SI{1}{\milli\meter}, \SI{1.5}{\milli\meter}, \SI{2}{\milli\meter}, \SI{2.5}{\milli\meter}, and \SI{3}{\milli\meter}) until it failed. Duct tape was applied to the 3D-printed step surface to increase friction with the robot's feet. The rough terrain experiment was conducted on a 3D-printed PLA surface made by generating a normal distribution of square prisms with a side length of \SI{5.3}{\milli\meter}, mean height of \SI{16}{\milli\meter}, height standard deviation of \SI{2.5}{\milli\meter}, and a maximum height of \SI{32}{\milli\meter}. We tested the robot using a motor voltage waveform with $f = $~\SI{8.3}{\hertz} and $A = $~\SI{3.2}{\volt}.

\paragraph{Energetic efficiency}
A significant advantage of passive dynamic walking at large scales is the resulting energy efficiency. For example, Ranger had a minimum cost of transport of only \num{0.19}, even lower than that of human walking \cite{bhounsule2014low}. Energy-efficient locomotion is even more important for small robots that have limited on-board energy storage in batteries.
We assess \robotname{}'s energetic efficiency by determining the robot's cost of transport (CoT). This efficiency metric is commonly used to evaluate a robot's power consumption during locomotion \cite{kuo2007choosing}. CoT is defined as follows: 
\begin{equation}
    \text{CoT} := \frac{\bar{P}}{mg\bar{v}}
\label{eqn:cot}
\end{equation}
\noindent where $\bar{P}$ is the average power consumed during walking, $mg$ is the weight of the robot, and $\bar{v}$ is the robot's average locomotion velocity. 
To measure the robot's power consumption during walking, we connected the robot's power input pins to a DC power supply (E36312A, Keysight) using thin enameled wires (50 AWG, MWS Wire Industries) with a current sensor in series (SEN0291, DFRobot). We then measured the voltage and current readings from the current sensor using a separate off-board microcontroller (MKR Zero, Arduino). To retain identical inertia parameters, all components were kept on-board although the battery was not connected. We then performed the same forward locomotion speed test at a waveform of $A = $~\SI{3.2}{\volt} and $f = $~\SI{8.3}{\hertz}, the parameters that achieved the fastest speed.

\section{Results}
\subsection{Effect of modifying existing design rules}

\begin{figure}[!tb]
\vspace{.5em}
\centerline{\includegraphics[width=\columnwidth]{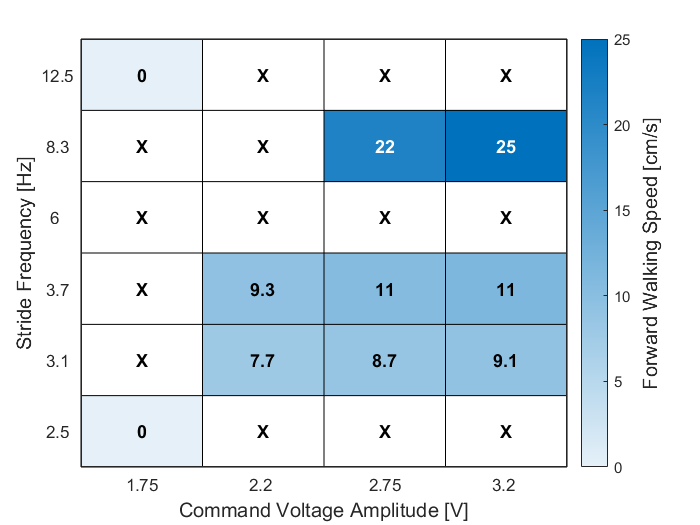}}
% \vspace{-.5em}
\caption{The robot's walking speed as a function of motor voltage amplitude and frequency. ``X" corresponds to trials where the robot turns randomly despite applying the same actuation signals. \SI{0}{\centi\meter/\second} walking speed corresponds to trials where the robot cannot make forward progress.}
\label{fig:sweep}
\end{figure}

\begin{figure*}[tb]
\centering
 \vspace{1em}
 \includegraphics[width=1.8\columnwidth]{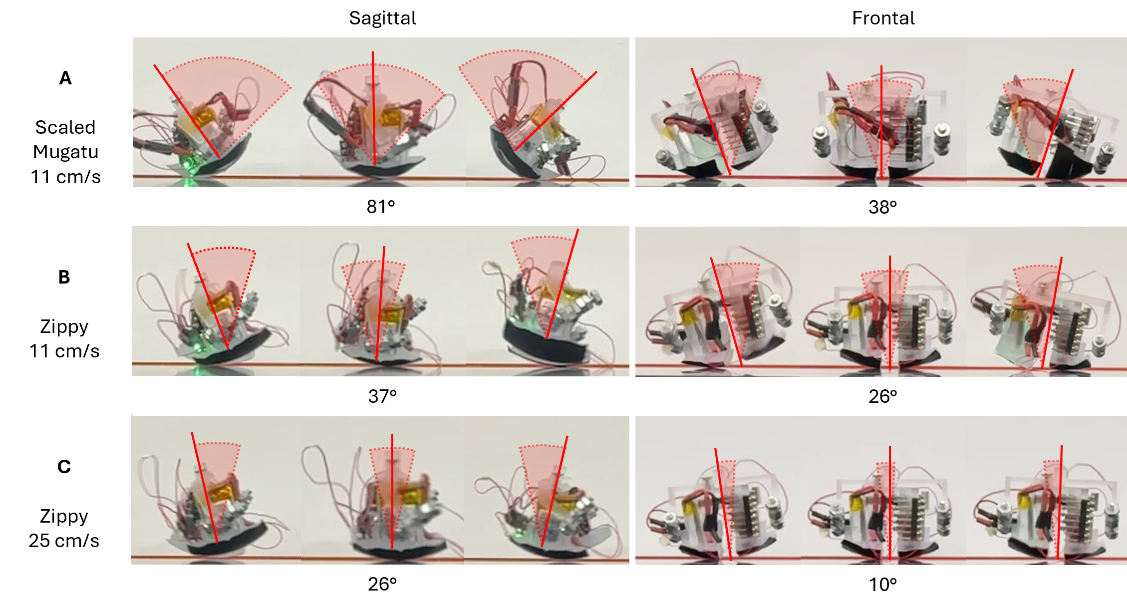}
 % \vspace{-1em}
\caption{Sagittal plane (pitch) and frontal plane (roll) oscillation amplitudes of: (A) Scaled Mugatu walking at \SI{11}{\centi\meter/\second},  (B) \robotname{} walking at \SI{11}{\centi\meter/\second} and (C) \robotname{} walking at \SI{25}{\centi\meter/\second}.}
\label{fig:rocking}
\end{figure*}

At the tested waveform parameters, \robotname{} and Scaled Mugatu were able to locomote at similar speeds over flat ground (approximately \SI{11}{\centi\meter/\second}). We captured high-speed video of the robots' motion and measured their hips' attitude over multiple frames to determine the maximum pitch and roll. As we compared both robots, clear differences in locomotion emerged.
% When the robot motion was captured from high-speed video (\Cref{fig:rocking}), clear differences in locomotion emerged.  
Scaled Mugatu walked with significant amounts of rolling and pitching (\Cref{fig:rocking}A), causing stumbling in the presence of small disturbances and tripping on its first steps. When the robot encountered minor perturbations caused by ground curvature or obstacles, the excessive pitching oscillation would often result in the robot falling forward or backward.

In \Cref{fig:rocking}B, it is clear that \robotname{}'s new design parameters significantly reduce excessive pitching (by \SI{44}{\degree}) in addition to reducing roll amplitude (by \SI{12}{\degree}). %The amount of pitch and roll amplitude during locomotion at the same speed are reduced by \SI{44}{\degree} and \SI{12}{\degree} respectively. %, we modified the feet to have an ellipsoidal profile, having a lower feet curvature in the sagittal plane than the frontal plane (\Cref{fig:dims}) as well as raising the center of curvature, which further reduced the feet curvature as we keep the leg length constant. 
We hypothesize that the ellipsoidal feet in particular help reduce maximum pitch while maintaining enough roll for foot clearance during leg swing. \Cref{fig:rocking}C shows that \robotname{}'s roll and pitch amplitudes do not increase even at higher locomotion speeds. %By modifying the feet shape to be an ellipsoid and reducing the forward pitching curvature, we reduced the amount of pitch and roll amplitude by \SI{44}{\degree} and \SI{12}{\degree} respectively (\Cref{fig:rocking}B). 
The ellipsoidal foot shape on \robotname{} was used for all further experiments.

\begin{figure*}[tb]
\centering
 \includegraphics[width=1.8\columnwidth]{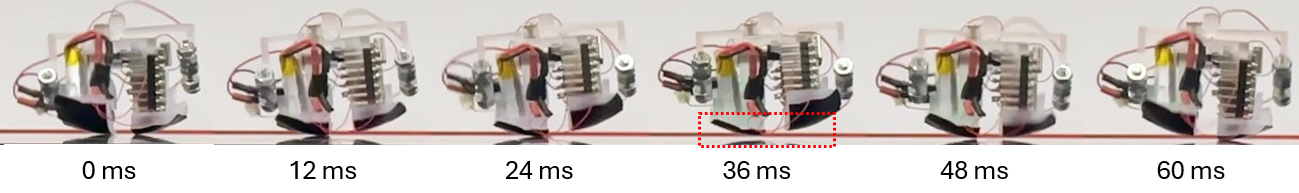}
 \vspace{-1em}
\caption{Key frontal frames of \robotname{} traversing at \SI{25}{\centi\meter/\second} displaying brief aerial phase when the swing leg impacts the hard stop (t = \SI{36}{\milli\second}).
}
\label{fig:skipping}
\end{figure*}

\subsection{Forward locomotion}
To determine the robot's fastest forward velocity and locomotion limits, we varied the stride frequencies and voltage amplitudes (\Cref{fig:sweep}). The robot's fastest speed of \SI{25}{\centi\meter/\second} used a waveform with $A = $~\SI{3.2}{\volt} and a $f = $~\SI{8.3}{\hertz}. At frequencies above \SI{8.3}{\hertz} and below \SI{3.1}{\hertz}, the robot produced excessive turning and chaotic steps. In some cases with amplitudes below \SI{2.2}{\volt}, the motor could not generate enough torque to fully lift the legs, causing the robot to step in place. 

We observed that the robot's forward velocity generally increases with the amplitude of the actuation waveform at any given frequency (\Cref{fig:sweep}). We also observed that as we increased actuation frequency within the range of walkable frequencies, the robot's speed increased at each given amplitude. 
However, there exists a range between stable walking frequencies where the robot cannot walk straight despite attempts to tune the motor voltage offset. We believe this instability occurs when the actuation frequencies are out of phase with the natural swinging and rolling dynamics of the robot, resulting in the hip actuation consistently assisting one foot's swing while hindering the other.

We attribute the drastic increase in speed as the stride frequency increased from \SI{3.7}{\hertz} to \SI{8.3}{\hertz} to the increase in average foot swing velocity coupled with an emergent skipping motion (\Cref{fig:skipping}). Because of the constant hip joint angle limit, at the same command voltage amplitudes, the hip motor takes the same duration to travel the entire range. At the \SI{3.7}{\hertz} stride frequency, the motor hits the hard stop early on (at \SI{10}{\percent} of the walking period) and the robot remains in the same pose while rolling on its feet until the motor voltage reverses (at \SI{50}{\percent} of the walking period) to swing the other leg. This means the robot relies on the passive rolling of the body to advance its position. At the \SI{8.3}{\hertz} stride frequency, the hard stop impact occurs at \SI{50}{\percent} of the walking period and causes the robot to skip off the ground briefly. This mode of locomotion was previously theorized in various models to have a wide range of speeds with a potential tradeoff between improved stability and reduced energy efficiency\cite{minetti1998biomechanics, asano2012limit}. This skipping gait causes the robot to land on the backward swinging leg each time, allowing the swinging motion to drive the robot forward instead, leading to a higher speed than walking \cite{andrada2016stability}.

\subsection{Turning}
\begin{figure}[tb]
\vspace{1em}
\centerline{\includegraphics[width=\columnwidth]{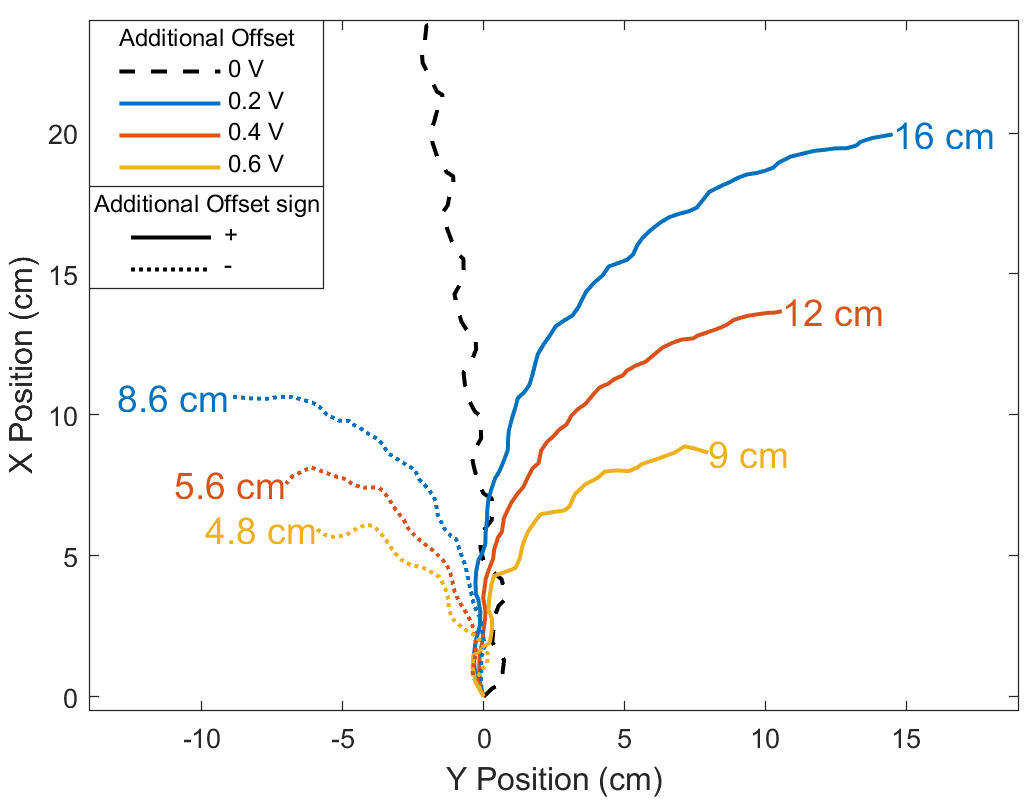}}
% \vspace{-1em}
\caption{The average robot turning trajectory at different radii of curvature when different additional DC offsets of the command voltage waveform are supplied (3 trials per turning radius). $V_0$ is the nominal offset used in tuning to achieve a straight walk.}
\label{fig:turning}
\end{figure}

The robot can be steered by varying the DC offset of the motor voltage \eqref{eqn:square_wave}. Changing the DC offset increases or decreases the flight time of each leg, which varies each leg's stride length, resulting in a curved walking path. We demonstrated the relationship between the turning radius and DC offset of the command voltage in \Cref{fig:turning}. As expected, when we increase the magnitude of the DC offset, we achieve tighter turns characterized by smaller turning radii. 

While the robot has a clear bias toward left turns (also visualized in \Cref{fig:rocking}B and \Cref{fig:rocking}C), we showed that the robot can execute turning maneuvers with a radius of curvature as low as but not limited to \SI{4.8}{\centi\meter} when given an additional \SI{0.6}{\volt} DC offset at a walking speed of \SI{8.8}{\centi\meter/\second}. When we reverse the offset, the robot can steer in the opposite direction with a \SI{9}{\centi\meter} radius of curvature.

\subsection{Efficiency}
We performed the cost of transport experiments near the peak recorded walking speed to observe how the energetic CoT of the robot varies with various actuator parameters. We obtained the minimum CoT of \num{11.2} at a walking speed of \SI{18.5}{\centi\meter/\second} with a waveform of $A = $~\SI{3.2}{\volt} and $f = $~\SI{8.3}{\hertz}. We attribute this decrease in speed to the tethered wiring setup that may have generated unexpected stiffness or friction during robot locomotion. We then decreased both frequency and amplitude to determine their relative effect on CoT. We measured: CoT = \num{13.0} ($f = $~\SI{8.3}{\hertz}, $A = $~\SI{2.75}{\volt}, and $\bar{v} = $~\SI{9.9}{\centi\meter/\second}); CoT = \num{22.6} ($f = $~\SI{3.7}{\hertz}, $A = $~\SI{3.2}{\volt}, and $\bar{v} = $~\SI{14.7}{\centi\meter/\second}).
To quantify how this energy efficiency can lead to longer runtimes, we allowed the robot to walk continuously at full speed until battery depletion, resulting in a measured runtime of \num{54} minutes.

\subsection{Step disturbance and rough terrain}
\begin{figure}[t]
\vspace{1em}
\centerline{\includegraphics[width=\columnwidth]{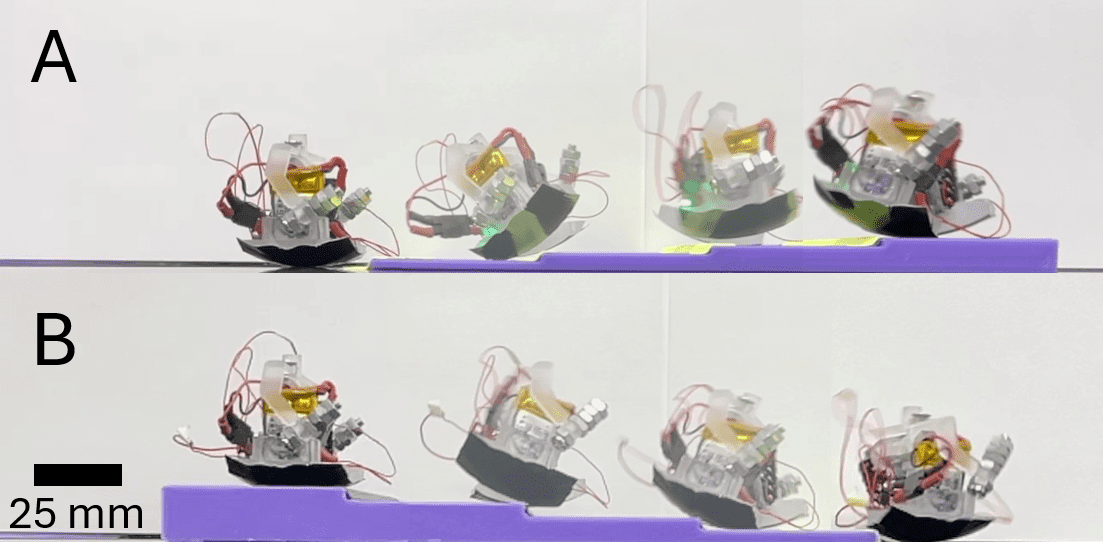}}
\caption{(A) The robot successfully walking up a flight of steps with a step height of \SI{2}{\milli\meter} and (B) stepping down a flight of steps with a step height of \SI{3}{\milli\meter} while remaining stable.}
\label{fig:steps}
\end{figure}

\begin{figure}[t]
\centerline{\includegraphics[width=\columnwidth]{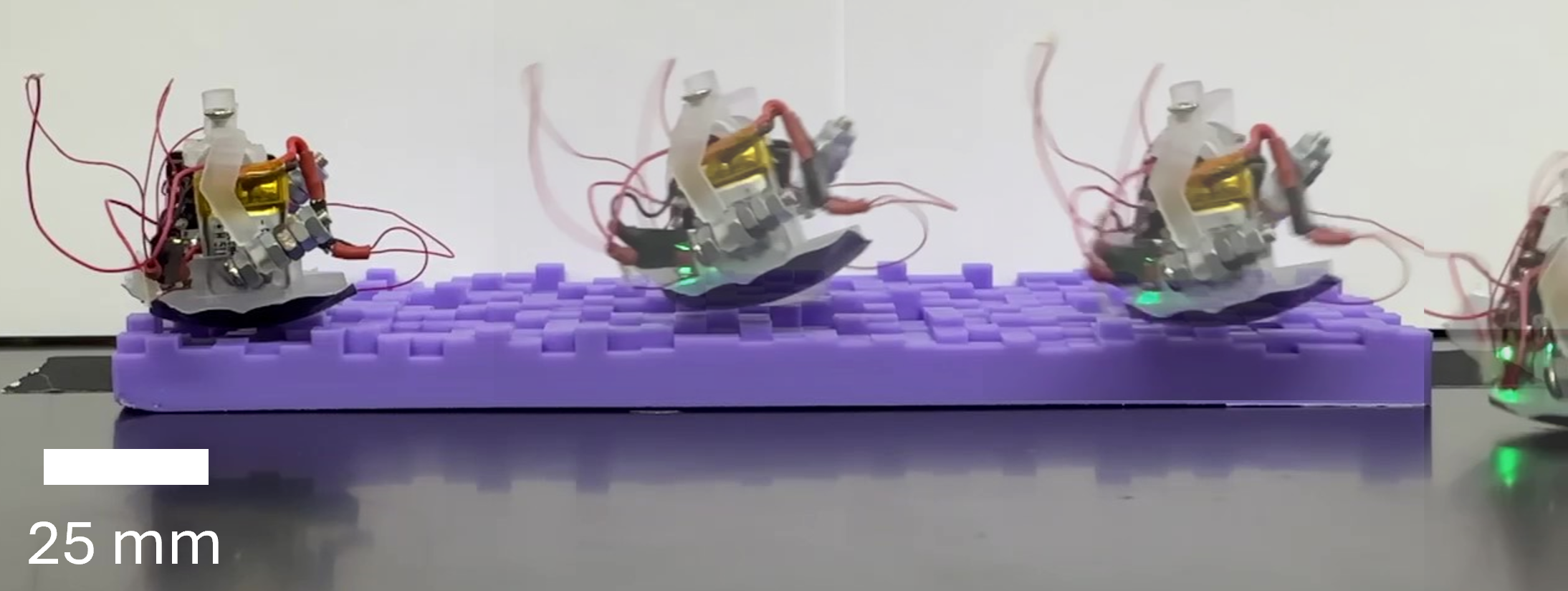}}
\caption{Key frames of the robot successfully walking across \SI{20}{\centi\meter} of a random obstacles terrain (\SI{5.3}{\milli\meter} by \SI{5.3}{\milli\meter} squares with a height standard deviation of \SI{2.5}{\milli\meter}) in \SI{1.2}{\second}.}
\label{fig:minefield}
\end{figure}

As seen in \Cref{fig:steps}, \robotname{} demonstrates a remarkable ability to ascend and descend small steps using only open-loop control and no actuation at the knee. The robot can climb \SI{2}{\milli\meter} (\SI{8}{\percent} leg length) steps and descend \SI{3}{\milli\meter} (\SI{12}{\percent} leg length) steps consistently without tipping over. \robotname{} also successfully traversed the 3D-printed rough terrain (\Cref{fig:minefield}) without tipping over at an average speed of \SI{16.7}{\centi\meter/\second}. We attribute the robot's ability to remain stable while traversing steps or uneven terrain to the relative size and ellipsoidal profile of the robot's feet, which limits forward and backward rocking when encountering terrain disturbances.

\section{Conclusions}
In this study, we presented ``\robotname{}", a \SI{3.6}{\centi\meter}-tall, self-contained bipedal walking robot that leverages principles from passive dynamic walking to achieve high-speed, stable locomotion with minimal actuation. \robotname{}'s design incorporates a single motor at the hip, ellipsoidal feet to enhance stability, and open-loop control, which collectively enable a forward speed of \num{10} leg lengths per second--the fastest among all bipedal robots at any scale. Additionally, \robotname{} can perform useful maneuvers such as turning and ascending steps, highlighting its ability to operate in rough terrain. This work demonstrates that scaling down passive dynamic walking principles can result in a compact, agile robot capable of stable locomotion.

Despite these achievements, several limitations were identified, particularly related to the robot's open-loop control strategy. The lack of sensor feedback control makes \robotname{} more susceptible to disturbances in yaw and necessitates precise parameter tuning for straight walking. Moreover, while \robotname{} can handle small steps and uneven terrain, its performance degrades in more challenging environments with larger obstacles. These limitations highlight the need for future enhancements such as leg compliance and sensor feedback to increase robustness and versatility.

Looking forward, we plan to integrate closed-loop feedback using an inertial measurement unit (IMU) already on the robot as part of the microcontroller board. Data from the IMU could enable real-time gait adjustments, thereby improving turning stability and robustness to disturbances. Furthermore, incorporating a microcontroller camera attachment could enable vision-based navigation and offline path planning. With these localization capabilities, multiple robots could be deployed in parallel to coordinate as a swarm for inspection and search and rescue operations in confined or hazardous environments.

% \section*{Acknowledgment}
% ACKNOWLEDGEMENT
\balance

\bibliographystyle{IEEEtran}
\bibliography{references}

\end{document}